\documentclass[10pt,twocolumn,letterpaper]{article}

\usepackage{cvpr}              

\usepackage[pagebackref,breaklinks,colorlinks,citecolor=cvprblue]{hyperref}
\usepackage{graphicx}
\usepackage{amsmath}
\usepackage{amssymb}
\usepackage{booktabs}
\usepackage{stfloats}
\usepackage{color}
\usepackage{times}
\usepackage[dvipsnames]{xcolor}

\newcommand{\PAR}[1]{\vskip3pt \noindent{\bf #1~}}
\definecolor{cvprblue}{rgb}{0.21,0.49,0.74}
\usepackage{overpic}
\usepackage{bm}
\usepackage{tabu}
\usepackage{tabularx}
\usepackage{bbding}
\usepackage{multicol}
\usepackage{multirow}
\usepackage{scalerel,stackengine}
\usepackage[accsupp]{axessibility}
\usepackage{array}
\usepackage[capitalize]{cleveref}
\crefname{section}{Sec.}{Secs.}
\Crefname{section}{Section}{Sections}
\Crefname{table}{Table}{Tables}
\crefname{table}{Tab.}{Tabs.}
\usepackage{algorithm}
\usepackage{algorithmic}
\usepackage[capitalize]{cleveref}
\crefname{section}{Sec.}{Secs.}
\Crefname{section}{Section}{Sections}
\Crefname{table}{Table}{Tables}
\crefname{table}{Tab.}{Tabs.}

\def\paperID{12443} 
\def\confName{CVPR}
\def\confYear{2024}

\title{
RELI11D: A Comprehensive Multimodal Human Motion Dataset and Method 
\vspace{-4mm}
}

\author{Ming Yan$^{1,2,3\ast}$\hspace{4mm} Yan Zhang$^{1,3\ast}$\hspace{4mm} Shuqiang Cai$^{1,3}$\hspace{4mm} Shuqi Fan$^{1,3}$\hspace{4mm} Xincheng Lin$^{1,3}$\hspace{4mm} Yudi Dai$^{1,3}$\hspace{4mm}\\ Siqi Shen$^{1,3\dagger}$\hspace{4mm} Chenglu Wen$^{1,3}$\hspace{4mm} Lan Xu$^4$\hspace{4mm} Yuexin Ma$^4$\hspace{4mm} Cheng Wang$^{1,3}$ \vspace{+1mm}\\
$^1$Fujian Key Laboratory of Sensing and Computing for Smart Cities, Xiamen University\\
$^2$National Institute for Data Science in Health and Medicine, Xiamen University\\
$^3$Key Laboratory of Multimedia Trusted Perception and Efficient Computing,\\ Ministry of Education of China, School of Informatics, Xiamen University\\
$^4$Shanghai Engineering Research Center of Intelligent Vision and Imaging, ShanghaiTech University\\
{}
\vspace{-18mm}
\and
\\
{}
}

\begin{document}
\twocolumn[{%
\renewcommand\twocolumn[1][!htb]{#1}%
\maketitle
\vspace{-9mm}
\begin{center}
    \centering
    \includegraphics[width=1\linewidth]{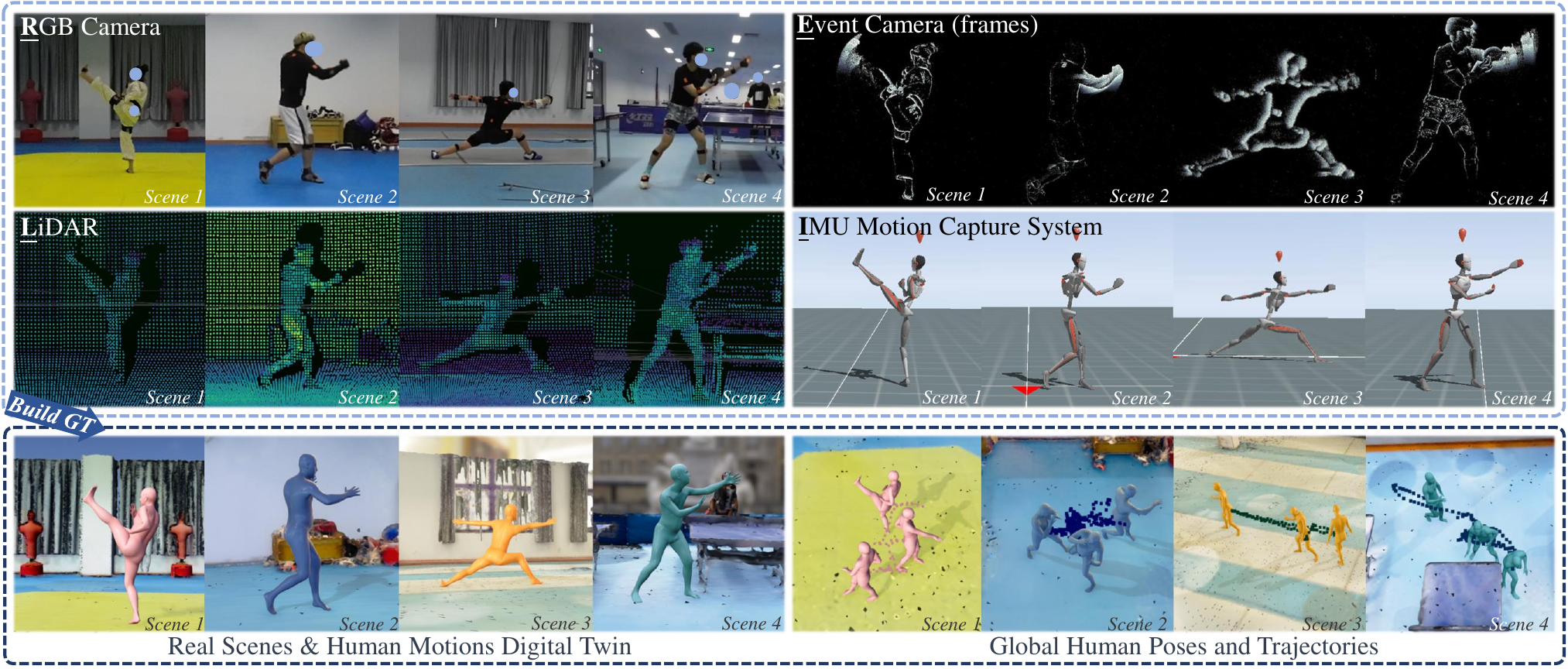}
    \vspace{-6mm}
    \captionof{figure}{RELI11D is a high-quality dataset that provides four different modalities and records movement actions(\textcolor[rgb]{0.56, 0.667, 0.863}{first two rows}). Our dataset's annotation pipeline can provide accurate global SMPL joints, poses as well as global human motion trajectories(\textcolor[rgb]{0.1254, 0.22, 0.392}{last row}).}

    \label{fig:gallery_reli}
\end{center}%
}]


\footnotetext{$^\ast$ Equal contribution.}
\footnotetext{$^\dagger$ Corresponding author.}

\begin{abstract}
\vspace{-4mm}

Comprehensive capturing of human motions requires both accurate captures of complex poses and precise localization of the human within scenes. Most of the HPE datasets and methods primarily rely on RGB, LiDAR, or IMU data. However, solely using these modalities or a combination of them may not be adequate for HPE, particularly for complex and fast movements. For holistic human motion understanding, we present \textbf{RELI11D}, a high-quality multimodal human motion dataset involves \textbf{L}iDAR, \textbf{I}MU system, \textbf{R}GB camera, and \textbf{E}vent camera. It records the motions of 10 actors performing 5 sports in 7 scenes, including 3.32 hours of synchronized LiDAR point clouds, IMU measurement data, RGB videos and Event steams. Through extensive experiments, we demonstrate that the RELI11D presents considerable challenges and opportunities as it contains many rapid and complex motions that require precise location. To address the challenge of integrating different modalities, we propose \textbf{LEIR}, a multimodal baseline that effectively utilizes LiDAR Point Cloud, Event stream, and RGB through our cross-attention fusion strategy. We show that LEIR exhibits promising results for rapid motions and daily motions and that utilizing the characteristics of multiple modalities can indeed improve HPE performance. Both the dataset and source code release publicly in \href{http://www.lidarhumanmotion.net/reli11d/}{http://www.lidarhumanmotion.net/reli11d/}, fostering collaboration and enabling further exploration in this field.

\end{abstract}

\vspace{-5mm}
\section{Introduction}
\label{sec:intro}

Human Pose Estimation (HPE)~\cite{Martinez17,alldieck2017optical,calabrese2019dhp19,Zhou16a,OpenPose,VIBE,SMPL2015,cai2022humman,shen2023x} is a challenging and long-standing research problem with significant potential for various applications, including AR/VR, autonomous driving, and sport analysis. Capturing complex and rapid human motions~\cite{chen2021sportscap} is particularly challenging, which requires the accurate estimation of poses and precise localization of individuals within various scenes.
Researchers adopt RGB cameras~\cite{SMPLERX, HMR,HUMOR_ICCV2021,survey2022,ICON,humanbench2023,HPERLICPR20} for HPE as they can capture appearance information, but they are light-sensitive and a monocular camera cannot provide depth information. Besides RGB imagery ~\cite{OSX,SPIN_ICCV2019,MonoPerfCap,DeepCap_CVPR2020,challencap,LiveCap2019tog}, there are multiple types of sensors that excel in capturing various aspects of human motions~\cite{survey2022}. RGBD sensors compensate for the absence of depth information, but their sensing range is limited. LiDAR~\cite{lidarcap} is light-insensitive and can provide 3D geometry. However, it suffers from sparsity and low frame rate issues. Inertial Measurement Units (IMUs)~\cite{DIP,PIP} is occlusion-free. Nevertheless, they should be body-worn and are subject to the drifting issue. Event cameras~\cite{eventcamera} can capture motions with high temporal resolution and dynamic range by measuring intensity change asynchronously. However, they do not provide appearance information. In conclusion, these sensors have distinct characteristics. Therefore, to obtain the holistic understanding of human motions, using multiple types of sensors is important.




HPE methods are partially driven by the development of human motion datasets~\cite{AMASS_ICCV2019}. Most of them use several types of sensors. As far as we know, there does not exist a human motion dataset that contains the RGB, LiDAR, Event, and IMU modalities. Such a multimodal dataset is beneficial for the community to better understand human motions. To this end, we introduce a multimodal human motion dataset, \textbf{RELI11D}, which involves four types of sensors: RGB cameras, LiDAR, Event cameras, and IMU measurements. It comprises data from 10 actors (2 females, 8 males), encompassing motions of 5 different sports (table tennis, taekwondo, boxing, fencing, and badminton) in 7 scenes. RELI11D includes a diverse range of synchronized data, consisting of 199.26 minutes of RGB videos, event streams, IMU motion capture data, and point cloud frames.

The rich modalities and annotations provided in our dataset enable benchmarking on a series of 3D HPE tasks. We quantitatively and qualitatively evaluate multiple state-of-the-art methods for these tasks. Most of these methods cannot deal with rapid, coherent, and complex movements that require precise location. The experimental results show that our dataset brings new challenges to current computer vision algorithms.


To address these challenges, we propose \textbf{LEIR}, a baseline that estimates global human poses using \textbf{L}iDAR point clouds, \textbf{E}vent streams, and \textbf{R}GB images. It effectively utilizes the geometry information from LiDAR, the motion dynamics encoded in events, and the appearance features in RGB images through our multimodal cross-attention bases method. The advantages of LEIR have been thoroughly validated through experiments. We show that leveraging multiple modalities is necessary for a comprehensive understanding of human motions. In summary, our contributions are listed below:

\begin{itemize}
    \item We present RELI11D, the \emph{first} HPE dataset consisting of the RGB, IMU, LiDAR, and Event modalities.
    \item  We provide a benchmark that enables the comparison of multiple methods using different modalities.
    \item We propose LEIR, a multi-modality baseline integrating the LiDAR point clouds, event streams, and RGB images for global human poses and trajectories estimation.
\end{itemize}

\label{sec:intro}

\section{Related Work}\label{sec:related}



\begin{table*}[!t]
     \vspace{-4mm}
     \centering
     \resizebox{1\linewidth}{!}{
     \begin{tabular}{lccccccccccc}
        \toprule[1pt]
        
        \multirow{2}{*}{Dataset} & \multicolumn{4}{c}{Sensor Modalities} & \multirow{2}{*}{Global} & \multirow{2}{*}{Frames} & \multirow{2}{*}{3D Scene} & \multirow{2}{*}{Motion} & \multirow{2}{*}{Real/} & \multirow{2}{*}{Number of} & \multirow{2}{*}{Number of}  \\

        \cline{2-5}
        
          & RGB & MoCap & LiDAR & Event  & Trajectory &  &  & &Synthetic & Sequences & Subjects   \\
        \midrule
        
         LiDARHuman26M\cite{lidarcap} &   \CheckmarkBold    &    IMU    &   \CheckmarkBold   &    -    &    -   & 184k &  - &   Daliy &   Real  & 20 & 13  \\
        HSC4D~\cite{HSC4D} &   -    &  IMU     &    \CheckmarkBold    &   -   &    \CheckmarkBold    &   10k   &  \CheckmarkBold & Daliy & Real & 8 & 1   \\
        SLOPER4D~\cite{dai2023sloper4d} &    \CheckmarkBold   &     IMU  &    \CheckmarkBold    &   -   &     \CheckmarkBold   &   100k   &  \CheckmarkBold & Daliy & Real  & 15 & 12   \\
        CIMI4D~\cite{yan2023cimi4d} &    \CheckmarkBold   &   IMU    &    \CheckmarkBold    &   -   &    \CheckmarkBold    &   180k   & \CheckmarkBold  & Climbing & Real  & 42 & 12  \\   
        LIPD~\cite{ren2023lidar} &   \CheckmarkBold    &   IMU    &    \CheckmarkBold    &   -   &    \CheckmarkBold    &  -    &  - & Obvious & Real  & 10 & 6   \\



        LiCamPose\cite{cong2022weakly} &   \CheckmarkBold    &   IMU    &    \CheckmarkBold    &   -   &    -    &  9k    &  - & Daliy & Real   & - & -   \\

        EMDB\cite{kaufmann2023emdb} &   \CheckmarkBold    &   EM    &    -    &   -   &    \CheckmarkBold    &  105k    &  - & Daliy & Real  & 81 & 10    \\

        SMART\cite{chen2021sportscap} &   \CheckmarkBold    &   -    &    -    &   -   &    -    &  110k    &  - & Sports & Real  & 640 & -   \\
        X-Avatar\cite{shen2023x} &   \CheckmarkBold    &   -    &    -    &   -   &    -    &  35k    &  \CheckmarkBold & Daliy & Real  & 233 & 20   \\

        BEHAVE~\cite{bhatnagar2022behave} &  \CheckmarkBold     &    -    &    -   &  -     &    \CheckmarkBold    &  15k &  -  & Interactions &   Real   & - & 8    \\
        
        RICH~\cite{RICH} &   \CheckmarkBold    &  -     &     -   &  -   &     \CheckmarkBold   & 577k & \CheckmarkBold  &  Interactions &  Real   & 142 & 22  \\
        AGORA~\cite{AGORA} &  \CheckmarkBold      &      - &    -    &   -  &     \CheckmarkBold   & 106.7K &  \CheckmarkBold &  Daliy &  Synthetic   & - & -  \\

        3D-FRONT HUMAN~\cite{yi2023mime} &    -   &   -    &    -    &  -   &   \CheckmarkBold     & - & \CheckmarkBold  &  Daliy &  Synthetic  & - & -   \\        
        BEDLAM~\cite{black2023bedlam} &  \CheckmarkBold     &     -  &   -     &  -   &     -   & 1M &  \CheckmarkBold &  Daliy &  Synthetic  & - & -  \\

        DHP19~\cite{calabrese2019dhp19} &  -     &     -  &   -     &  \CheckmarkBold   &     -   & 87k &  - &  Daliy &  Real  & - & 17  \\

        MMHPSD~\cite{zou2021eventhpe} &  \CheckmarkBold     &     -  &   -     &  \CheckmarkBold   &     -   & 240k &  - &  Daliy &  Real  & 84 & 15  \\
        
        \midrule
        \textcolor[rgb]{0.066,0.466,0.69}{\textbf{RELI11D(Ours)}}
         &   \textcolor[rgb]{0.066,0.466,0.69}\CheckmarkBold    &   \textcolor[rgb]{0.066,0.466,0.69}\CheckmarkBold    &    \textcolor[rgb]{0.066,0.466,0.69}\CheckmarkBold    &  \textcolor[rgb]{0.066,0.466,0.69}\CheckmarkBold   &      \textcolor[rgb]{0.066,0.466,0.69}\CheckmarkBold  & \textcolor[rgb]{0.066,0.466,0.69}{\textbf{239k}} &  \textcolor[rgb]{0.066,0.466,0.69}\CheckmarkBold \textcolor[rgb]{0.066,0.466,0.69}{\textbf{(7)}} &  \textcolor[rgb]{0.066,0.466,0.69}{\textbf{Sports}} & 
         \textcolor[rgb]{0.066,0.466,0.69}{\textbf{Real}} & 
         \textcolor[rgb]{0.066,0.466,0.69}{\textbf{48}} & 
         \textcolor[rgb]{0.066,0.466,0.69}{\textbf{10}} \\

        \bottomrule[1pt]
        \end{tabular}%
    }
    \vspace{-2mm}
        \caption{Comparisons with related datasets. The "-" symbol indicates that it is not included in the dataset.
}
        \vspace{-4mm}
        \label{tab:data_compare}

 \end{table*}

\subsection{Single modality Datasets and Methods}

Many RGB motion datasets~\cite{PROX, cai2022humman, shao2022diffustereo,cheng2023dna} are collected using marker-based (e.g., Human3.6M~\cite{Ionescu2014Human36MLS}, HumanEva\cite{Sigal2009HumanEvaSV}) or marker-less methods (e.g., MPI-INF-3DHP\cite{Mehta2017Monocular3H}).  RGB-Based methods have flourished in recent years with a variety of approaches~\cite{chen2021sportscap,Bogo2016KeepIS,SIMPX, MAED, pifu, pifuhd, ICON, wang2023nemo, GFPose2023, MotionBert2023, TORE2023, PARE_ICCV2021, HUMOR_ICCV2021, Su2020RobustFusionHV,yang2021s3,Patel2020TailorNetPC}, but most still focus solely on single RGB modality construction, with only a few considering the estimation of global trajectories~\cite{li2021hybrik, sun2023trace, li2023niki,GLAMR}.

LiDAR can directly acquire three-dimensional spatial information. P4T~\cite{Fan2021Point4T} and STCrowd~\cite{cong2022stcrowd} use LiDAR point clouds to segment the human body. LiDARCap~\cite{lidarcap} performs 3D HPE through LiDAR point clouds. 

An event camera generates a stream of events. Each pixel of the event camera responds asynchronously and independently to illumination change and generates an event if the change exceeds a threshold, which makes event cameras excel at capturing the local motions of objects~\cite{TOREPAMI23,E2PNet}. DHP19~\cite{calabrese2019dhp19} and \cite{lifting2021} perform 2D HPE by treating event streams as image frames. EventCap~\cite{EventCap} estimates 3D poses through a monocular event camera. EventHPE~\cite{EventHPEICCV21} performs HPE through a flow-based approach. EventPointPose~\cite{EventPointPose3DV22} regresses poses through event point clouds.

IMU-based methods ~\cite{pons2011outdoor,Yi2021TransPoseR3,SIP,DIP} are environment-independent and occlusion-free. ~\cite{pan2023fusing} fuses visual and inertial information for HPE; EgoLocate ~\cite{yi2023egolocate} estimates human poses based on sparse IMUs; ~\cite{rempe2020contact} estimates poses based on physical contact. However, their necessity to be worn poses practical challenges and difficulties.

\subsection{Multi-modality Datasets and Methods}


TotalCapture\cite{Trumble2017TotalC3} collects human poses in a studio through multi-view cameras and IMUs. 3DPW~\cite{3DPW} collects subjects walking in a city through IMU and a hand-held camera. 
PedX\cite{Kim2019PedXBD} records pedestrian poses through stereo images and LiDAR point clouds. ImmFusion~\cite{chen2023immfusion}, FusionPose~\cite{cong2022weakly}, and \cite{zheng2022multi} use RGB and LiDAR body point clouds to reconstruct human poses. LIP~\cite{ren2023lidar} reconstructs poses using sparse IMUs with LiDAR. 
Besides using a monocular camera and IMUs, LiDARHuman26M~\cite{lidarcap}, HSC4D~\cite{HSC4D}, SLOPER4D~\cite{dai2023sloper4d}, CIMI4D~\cite{yan2023cimi4d} collect human poses through a static monocular LiDAR, a body-mounted LiDAR, a head-mounted LiDAR, and a LiDAR respectively. 
Researchers have also explored some other sensors for human motions, such as WIFI~\cite{WiPoseMobiCom20,WinectRen2021,MetaFiIOJ2023},mmWave~\cite{an2022mri,chen2022mmbody,chen2023immfusion}, and electromagnetic sensor~\cite{kaufmann2023emdb}.

\label{sec:relatedWork}

\section{RELI11D: a multimodal motion dataset}\label{sec:dataset}
\begin{figure}[tb]
    \centering
    \includegraphics[width=1\linewidth]{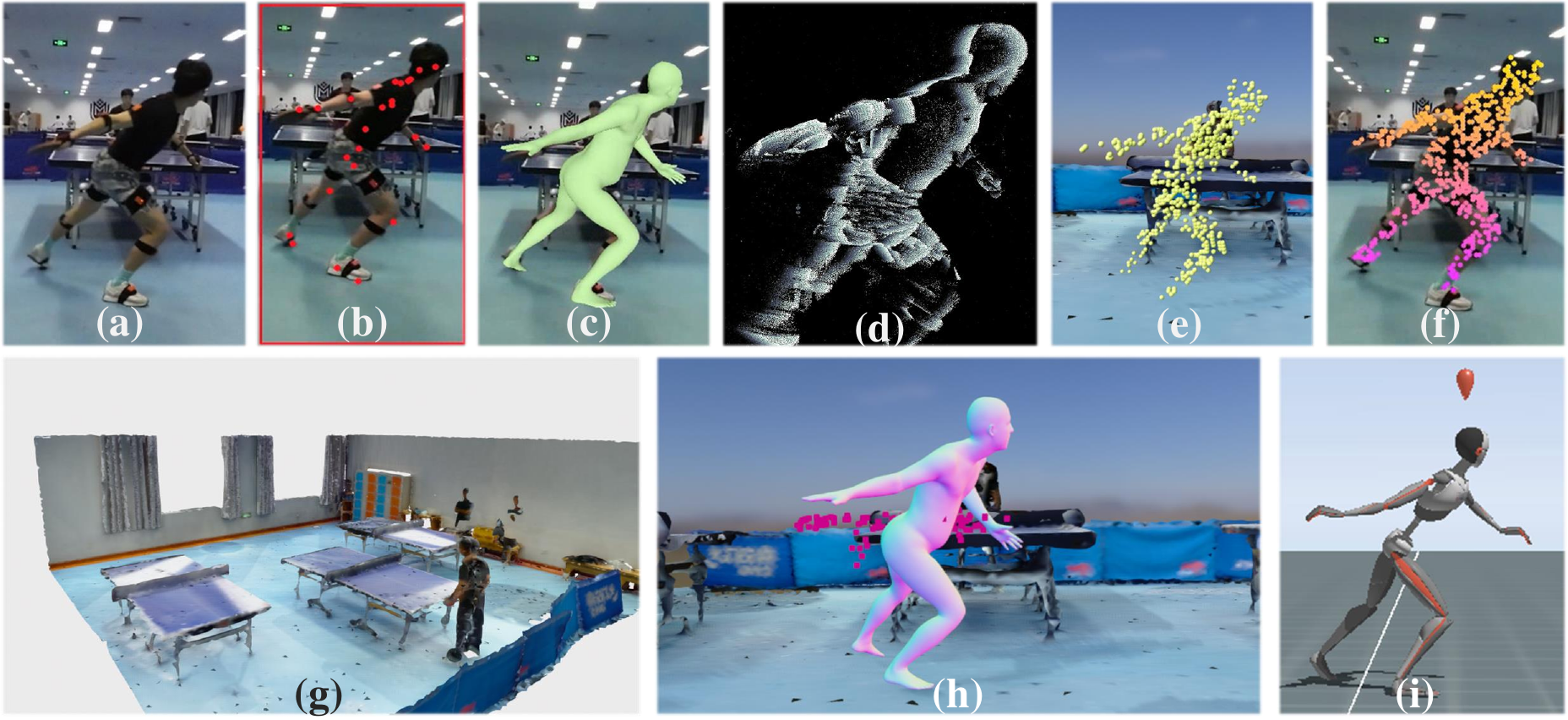}
    \vspace{-6mm}
    \caption{RELI11D provides rich data and annotations: (a) RGB Videos, (b) 2D Annotation, (c) 2D SMPL Poses, (d) Events, (e) 3D Point Clouds, (f) 2D Point Clouds, (g) High Precision Scene Meshes, (h) 3D SMPL Shape, Poses, and Trajectories, (i) IMUs Measurements.}
    \vspace{-2mm}
    \label{fig:multimodality}
 \end{figure}

\vspace{-1mm}
RELI11D is a multimodal high-quality human movement dataset that contains five different categories of sports: table tennis, taekwondo (examination movements and free exercises), boxing (traditional boxing, free fighting, Muay Thai), fencing (saber, epee, foil), and badminton. It collects 4 modalities for a total of 48 sequences of 199.2 minutes (3.32 hours) of synchronized RGB camera video, Event camera streams, IMU Measurements Data, and LiDAR point clouds. In total, there are 239k frames of human body point clouds. In RELI11D, we invite 10 volunteers to collect sports in 7 scenes. All the participants agree that their recorded data may be used for scientific purposes. \cref{fig:multimodality} describes the rich modalities and annotations the community can get from RELI11D.
\cref{tab:data_compare} provides statistics comparing with other publicly available human pose datasets. As far as we know, RELI11D is the \emph{first} dataset that consists of RGB, LiDAR, IMU, and Event modalities. Moreover, it contains high-precision 3D scans of 7 scenes , global poses and trajectories of each actor, contributing to comprehensive human scene perception. 

 \begin{figure}[tb]
    \centering
    \includegraphics[width=1\linewidth]{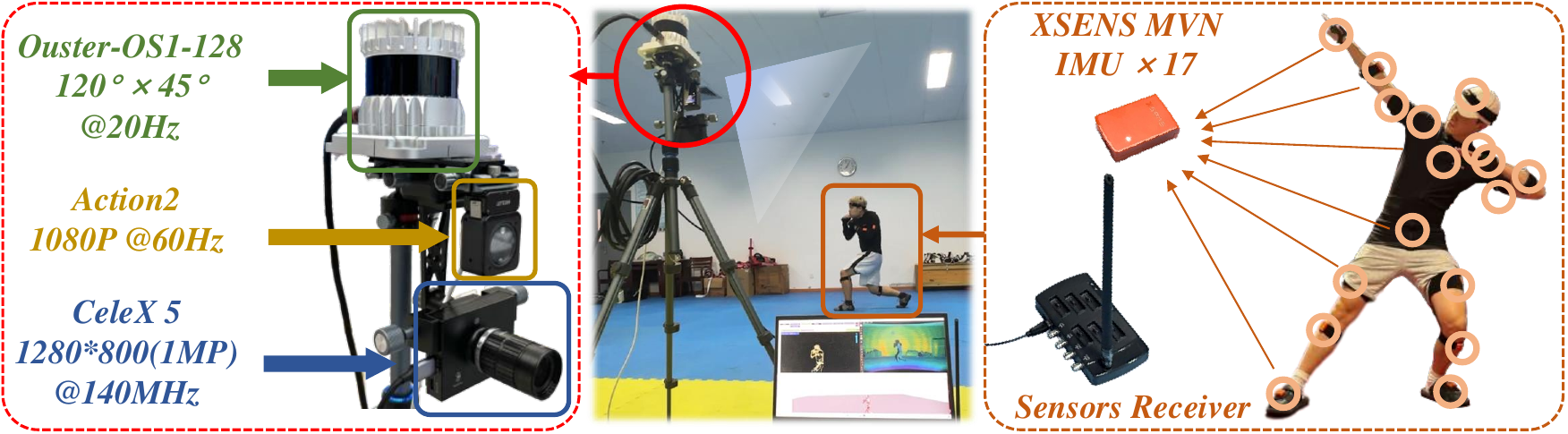}
    \vspace{-6mm}
    \caption{Portable Human Motion Capturing system.}
    \vspace{-5mm}
    \label{fig:Hardware}
 \end{figure}

 \begin{figure*}[tb]
    \centering
    \includegraphics[width=1\linewidth]{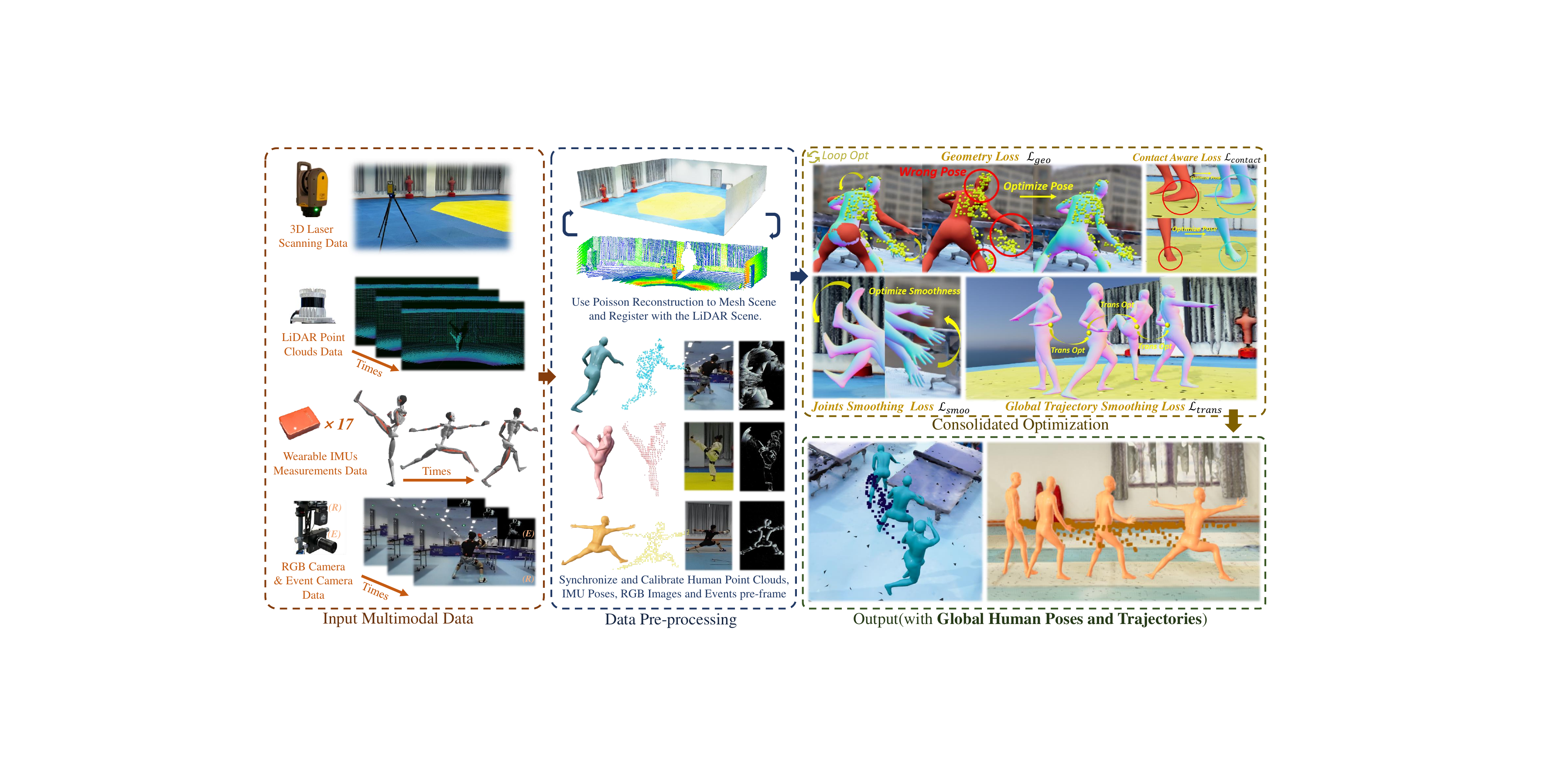}
    \vspace{-7mm}
    \caption{\textbf{Overview of main annotation pipeline.} The dotted boxes of different colors represent different data processing stages, and the arrows represent the data flow direction. \textbf{Dotted box:} The input of each scene sequence consists of RGB videos, point cloud sequences, IMU measurements, events flow(times axis), and 3D laser scanning data. The data pre-processing stage calibrates and synchronizes different modalities. The consolidated optimization includes the global pose and translation based on multiple constraint losses.}
    \vspace{-4mm}
    \label{fig:dataPipeline}
 \end{figure*}
 \vspace{-1mm}
\subsection{Hardware and Configuration}
 \vspace{-1mm}
\label{sec:Hardware}
The motion collection device is composed of multiple sensors that can collect motions indoor and outdoor. As shown in~\cref{fig:Hardware}, we use LiDAR (Ouster-OS1, 128-beam) to capture 3D dynamic point clouds at 20 frames-per-second (FPS), a monocular RGB camera (DJI Action 2, 4096×3072) to record RGB video at 60 FPS, and an event camera (CeleX-V~\cite{CeleX-V}, 1280x800) to record event streams. For each sports scene, we use the Trimble X7 3D laser scanning system to reconstruct a high-precision RGB 3D point cloud for it, totaling 80 million points for each scene. 



Each volunteer wears an Xsens MVN inertial motion capture system. It contains 17 IMUs, which record poses at a speed of 60 FPS. To obtain their body shape (SMPL $\beta$), we scan their body using a handheld point cloud modeling equipment and obtain $\beta$ through IPNet~\cite{bhatnagar2020combining}.

\PAR{Human Pose Model and Label.} A human motion is denoted by $M=(T, \theta, \beta)$, where $T$ represents the $N\times3$ translation parameters, $N\times\theta$ is the $24\times3$ pose parameter, and $\beta$ is the $10$ dimensions shape parameter following SMPL~\cite{SMPL2015}, $N$ is the frame count. As IMUs suffer severe drifting for long-period capturing, we seek to find the precise $T$ and $\theta$ for RELI11D as annotation labels. 



\vspace{-1mm}
\subsection{Data Annotation Pipeline}
\label{sec:DataPipeline}

The data annotation pipeline consists of 3 stages: pre-processing, consolidated optimization, and manual annotation. Fig.~\ref{fig:dataPipeline} depicts the main annotation pipeline. 

\vspace{-2mm}
\subsubsection{Multimodal Data Pre-processing Stage}
\vspace{-1mm}
\label{Sec:Data Preprocessing}
\PAR{Scene reconstruction.} For each RGB high-precision static 3D point cloud scene, we convert it into a mesh scene composed of triangular patches through implicit surface Poisson reconstruction~\cite{kazhdan2013screened}. Data in this format can more accurately calculate the interaction between the human body and the environment is convenient.



\PAR{Time synchronization.}
Synchronization between the IMU, LiDAR, RGB video, and event streams is achieved by detecting spikes in jump events. In each motion sequence, the subject jumps in place, and we design a peak detection algorithm to automatically find the height peaks in the IMU and LiDAR trajectories. RGB videos and IMU data are downsampled to 20/fps, consistent with LiDAR’s frame rate. The event stream is divided into multiple event frames $\mathbb{E}=\{\mathbb{E}_{t_i}\}^N_{i=1}$. $\mathbb{E}_{t_i}$ is the set of events whose time stamp $t$ satisfies $t_{i-1} <t\le t_{i-1}$.

\PAR{Calibration.}
Initially, we register the LiDAR sparse point scene and the high-precision scene of each sequence to the same coordinate system. Next, for each frame, we isolate the human body point clouds, and derive the human poses based on it. The movement sequence of a person in world coordinates \{$W$\} is represented by $\bm{M^W}= (T^W, \theta^W, \beta)$. $T^I$ and $\theta^I$ in $M^I = (T^I, \theta^I, \beta)$ are provided by IMUs. $\theta^W =R_{WI} \theta^I$ is used as the initial poses, where $R_{WI}$ is a rough calibration matrix from the IMU coordinate system to the world coordinate system. As the translation measured by IMUs is not accurate~\cite{yan2023cimi4d}, we use the position of the center of the human hip in the point clouds as $T^W$. Lastly, we execute frame-level temporal synchronization and spatial calibration for the scenes and all the modalities. 

\vspace{-2mm}
\subsubsection{Consolidated Optimization}
\vspace{-1mm}
\label{Sec:Blending optimization}

We utilize contact aware loss $\mathcal{L}_{contact}$, smoothness loss $\mathcal{L}_{smoo}$ and geometry loss $\mathcal{L}_{geo}$ to perform consolidated optimization of global poses and trajectories to obtain accurate and scene-natural human motion. Please refer to the supplementary for a detailed formulation of these losses. We minimize the overall loss which is defined as follows.
\begin{equation}
	\begin{split}
    \mathcal{L}=
        \lambda_{c}\mathcal{L}_{contact} + \lambda_{s}{\mathcal{L}}_{smoo} +\lambda_{g}{\mathcal{L}}_{geo} 
    \end{split}
\end{equation}
\noindent
where $\lambda_{c},\ \lambda_{s},\ \lambda_{g}$ are loss coefficients. 

\PAR{Contact Aware Loss.} The $\mathcal{L}_{contact}$ term combines scene constraints $\mathcal{L}_{sceneC}$ and self-penetration constraints $\mathcal{L}_{selfC}$ to improve the quality of local human poses. First, we use $\lambda_{sceneC}$ to penalize the vertices in the human SMPL mesh that penetrate the scene mesh. To avoid the self-penetration problem, we use the self-penetration constraint $\mathcal{L}_{selfC}$.

\PAR{Smoothness Loss.} We introduce $\mathcal{L}_{smoo}$ to ensure the motion smoothness. It includes (1) the global trajectory smoothing term $\mathcal{L}_{trans}$, which smooths the human body motion by minimizing pelvis acceleration. (2) The body posture smoothing term $\mathcal{L}_{poses}$ maintains the stability of the entire human body motion by minimizing the angular velocity of each pelvis-related joint. (3) The human joints smoothing term $\mathcal{L}_{joints}$ smooths SMPL joint acceleration. 

\PAR{Geometry Loss.} 
Point clouds contain the geometry of human motion, we use $\mathcal{L}_{geo}$ to make visible SMPL vertices approximate the geometric relationship. Following \cite{HSC4D}, for each SMPL mesh, we use \cite{katz2007direct} to remove invisible mesh vertices from the LiDAR perspective. $\mathcal{L}_{geo}$ is the 3D Chamfer distance between body point and visible SMPL vertices.

\vspace{-3mm}

 \begin{figure*}[!tb]
    \centering
    \includegraphics[width=1\linewidth]{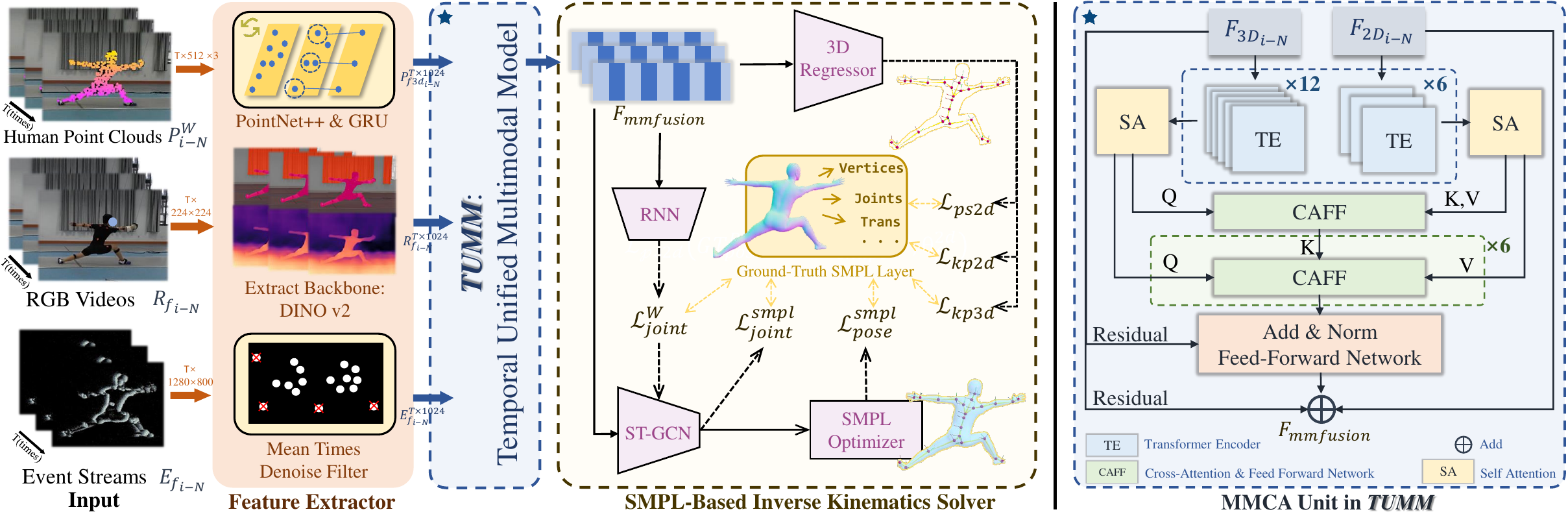}
    \vspace{-7mm}
    \caption{\textbf{Overview of LEIR method (Left) and Multimodal Cross-Attention Unit (Right).} \textcolor[rgb]{0.77, 0.35, 0.066}{Orange arrows} represent different modalities of data input. \textcolor[rgb]{0.184, 0.333, 0.592}{Dark blue arrows} represent the inputs and outputs data flows of the TUMM model. Dotted arrows represent the predicted data and calculation loss with ground truth.} 
    \vspace{-4mm}
    \label{fig:BaselineFramework}
 \end{figure*}

\vspace{-1mm}
\subsubsection{Manual Annotation and Verification Stage}
\vspace{-1mm}
\label{sec:manual_annotation}


We manually correct the pose and translation parameters of a subject's motions for some artifacts. Further, an external person has examined our dataset. We have adjusted the imprecise annotations pointed out by this person.



\label{sec:ConstructingData}

\section{LEIR: A multimodal HPE baseline}
We propose LEIR, a multi-modality baseline for human motion estimation. Given the synchronized LiDAR point clouds, RGB images, and event streams that are captured by multiple sensors, the task of the baseline is to predict the 3D pose of the human in the world coordinate system. 

Many existing two-modal-based methods~\cite{girdhar2023imagebind, kim2021vilt, li2021align} adopt a two-tower architecture, where each tower processes only one modality. LEIR aims to fully model three (rather than two) modalities, which is non-trivial because each modality contains different information.

As is depicted in \cref{fig:BaselineFramework}, LEIR consists of three major modules: feature extractors, the temporal unified multimodal model (TUMM), and SMPL-based inverse kinematics solver. For each modality, feature extractors are used to extract its features, which are fused through the TUMM modules to fully utilize the 3D geometric information of point clouds, the appearance information of RGB images, and the temporal dynamics of event streams. In the end, the fused features are fed into SMPL solver to obtain the estimated poses. Please see the appendix for more details.




\subsection{Feature Extraction}
\vspace{-1mm}

\PAR{RGB Feature Extraction.}
For each RGB frame, we specifically target the human body by applying a bounding box and extracting its corresponding feature $R_{{f}_{i-N}}$ using an RGB encoder (DINOv2~\cite{oquab2023dinov2}).
 
\PAR{LiDAR Feature Extraction.}
For the human body point clouds $P_{i-N}^W$, we extract its features $P_{{f3d}_{i-N}}$ by feeding the point clouds into a PointNet++\cite{qi2017pointnet++} and a GRU network.

\PAR{Event Feature Extraction.}  For each event frame $\mathbb{E}_{t_i}$, we employ an average time sampling filter with adjacent point denoising~\cite{yan2022lightweight} to effectively process the noise in the frame. This filtering technique enhances the visibility of changes in human body movement between frames. Subsequently, we aggregate all the events in a frame based on their pixel location and polarity, thus generating an image-like event frame. The features $E_{{f}_{i-N}}$ of this frame are then extracted using an RGB encoder (DINOv2).

\subsection{Temporal unified multimodal model (TUMM)}

 Previous methods, such as \cite{zheng2022multi}, fuse the LiDAR and RGB modalities through projection rely on accurate calibration, which may not always available. Moreover, it is unclear how to effectively fuse LiDAR and Event modalities. 


To automatically learn correspondence among three modalities and eliminate calibration sensitivity, TUMM uses the cross-attention strategy. The TUMM module consists of two steps. In the first step, the LiDAR point clouds and the RGB images are fused using the multimodal cross-attention unit (MMCA), LiDAR point clouds and the event frames are fused using MMCA as well. This step aims to effectively integrate the geometry information with appearance information, and integrate the geometry information with motion dynamics. In the second step, the features obtained from the first step are further fused using MMCA, which allows a comprehensive integration of the features from different modalities.

The design of the MMCA unit is depicted in the right part of ~\cref{fig:BaselineFramework}. The LiDAR features $F_{{3D}_{i-N}}$ and the RGB/event features $F_{{2D}_{i-N}}$ are processed through a series of transformer encoders \cite{vaswani2017attention} and self-attention mechanisms. MMCA employs a 2-layer cross-attention structure, using the fused keys as intermediaries to match and align two sources of information. In the first layer, the features from the LiDAR act as queries, while the features from the RGB/events serve as keys and values. In the second layer, the output from the last layer serves as the keys; the LiDAR feature and RGB feature serve as query and value, respectively. The output features are obtained by element-wise addition of the input features and the results of the cross-attention structures. For the second step of TUMM, the 2D (right) branch of MMCA is replaced by a 3D branch. The output results of the previous step are fed into MMCA for finding correspondence among three modalities.

\subsection{SMPL-Based inverse motion solver} 

The fused features $F_{mmfusion}$ are utilized in three branches within the network. In the first branch, these features are fed into a 3D regressor to estimate 3D joints and camera intrinsic parameters. This branch involves three different loss functions. $\mathcal{L}_{ps2d}$ serves as a projection loss, ensuring that the 2D appearance of the SMPL model aligns with the human body in pixel coordinates. $\mathcal{L}_{kp2d}$ and $\mathcal{L}_{kp3d}$ are used to respectively constrain the 2D and 3D joints of the human body. In the second branch, the features are fed into an RNN network that predicts 3D human joints in the world coordinate system. $\mathcal{L}_{joint}^{W}$ is employed to encourage alignment with the labels. The third branch employs an ST-GCN ~\cite{yan2018spatial}, where the fused features are used to predict 3D human joints. In addition, we apply $\mathcal{L}_{joint}^{smpl}$ to ensure accurate joint orientation. Finally, the outputs of the third branch are passed through an SMPL optimizer to obtain the human pose in axis-angle form, and $\mathcal{L}_{pose}^{smpl}$ is employed to enforce alignment with the ground truth poses. The overall loss function used in LEIR combines all these individual losses, which is defined as follows.

\vspace{-4mm}

\begin{equation}
	\begin{split}
        \mathcal{L}_{ps2d} +
        \mathcal{L}_{kp2d} +
        \mathcal{L}_{kp3d} +
        \mathcal{L}_{joint}^{W} +
        \mathcal{L}_{pose}^{smpl} +
        \mathcal{L}_{joint}^{smpl}
    \end{split}
\end{equation}

To evaluate the performance of different input modalities using the same network, we employ the same training strategy as~\cite{zhang2023meta}, which freezes unnecessary model parameters based on different input modalities, thus enabling multimodal or single-modal training and inference.

\section{Experimental Results}

\PAR{Evaluation metrics.} We report Mean Per Joint Position Error (MPJPE), Procrustes Aligned Mean Per Joint Position Error (PA-MPJPE), Percentage of Correct Keypoints (PCK0.3), Per Vertex Error (PVE), Acceleration Error($mm/s^2$) (ACCEL). Regarding the evaluation of trajectory errors, we use Global MPJPE (GMPJPE) to calculate the mean per joint position error of the SMPL model in global coordinates, global human body root node Translation Error(T-Error). The PCK0.3 is calculated as a percentage, while other indicators are in $mm$.


\subsection{Dataset Evaluations}

\begin{table}[tb]
	\centering
    \resizebox{\linewidth}{!}{
    \begin{tabular}{cccccc}
     \toprule
     Sport sequences  &  ACCEL$\downarrow$ & MPJPE$\downarrow$  & PA-MPJPE$\downarrow$ & PVE$\downarrow$ & PCK0.3$\uparrow$ \\
     \midrule
    \multirow{1}*{Pingpong}  &0.72/3.70 & 22.31/58.42 & 21.90/58.62 & 31.89/85.55  & 0.98  \\

    \multirow{1}*{Badminton} & 0.84/1.97 & 27.91/67.57 & 26.19/64.11 & 28.45/63.20  & 0.95 \\

    \multirow{1}*{Taekwondo} & 1.25/2.89 & 29.28/68.23 & 25.02/63.13 & 36.83/66.23  & 0.95 \\

    \multirow{1}*{Boxing} & 1.79/5.22 & 31.42/67.56 & 26.43/50.83 & 38.45/75.45  & 0.96 \\

    \multirow{1}*{Fencing} & 0.40/5.00 & 25.66/62.73 & 20.54/52.09 & 27.73/53.68  & 0.98 \\
     \bottomrule
     \end{tabular}%
     }
     \vspace{-3mm}
        \caption{Quality of the optimization process.  Each cell reports the mean and maximal error metrics which are separated by ``/''.} %
     \vspace{-2mm}
	\label{tab:eva_annotation}
\end{table}

\begin{table}[tb]
    \centering
	\resizebox{\linewidth}{!}{
    \begin{tabular}{ccc|ccc|ccc}
    \toprule
    \multicolumn{3}{c|}{Constraint term} & \multicolumn{3}{c|}{PingPong1} & \multicolumn{3}{c}{Boxing1} \\
    
    $\mathcal{L}_{contact}$ & $\mathcal{L}_{smoo}$ & $\mathcal{L}_{geo}$ & ACCEL$\downarrow$ & MPJPE$\downarrow$ & PA-MPJPE$\downarrow$ & ACCEL$\downarrow$ & MPJPE$\downarrow$  & PA-MPJPE$\downarrow$   \\
    \midrule
    \textcolor{red}{\XSolidBrush} & \textcolor{red}{\XSolidBrush} &  \textcolor{red}{\XSolidBrush} & 5.29 & 29.14 & 20.7 & 6.79  & 31.20 & 25.65  \\
    \textcolor{green}{\Checkmark} & \textcolor{green}{\Checkmark} &  \textcolor{red}{\XSolidBrush} & 1.96 & 13.75 & 10.11 & 1.52  & 11.89 & 14.96  \\
    \textcolor{green}{\Checkmark} & \textcolor{red}{\XSolidBrush} &  \textcolor{green}{\Checkmark} & 4.03 & 15.19 & 12.91 &  5.50   & 18.82 & 12.13   \\
    \textcolor{red}{\XSolidBrush} & \textcolor{green}{\Checkmark} &  \textcolor{green}{\Checkmark} & 1.29 & 18.14 & 13.80 &  1.10  & 12.23 & 9.66  \\
    \midrule
    \textcolor{green}{\Checkmark} & \textcolor{green}{\Checkmark} &  \textcolor{green}{\Checkmark} & \textbf{0.92} & \textbf{12.25} & \textbf{9.31} & \textbf{0.85}   & \textbf{10.26} & \textbf{8.35}  \\
    
    \bottomrule
    \end{tabular}%
    }
    \vspace{-3mm}
    \caption{Evaluation of Consolidated Optimization for different constraints. Unit: $mm$}
    \vspace{-4mm}
    \label{tab:eva_RELI}
\end{table}

    

    

\label{sec: Dataset Evaluations}
\begin{figure}[!tb]
    \vspace{+2mm}
    \centering
     \includegraphics[width=1\linewidth]{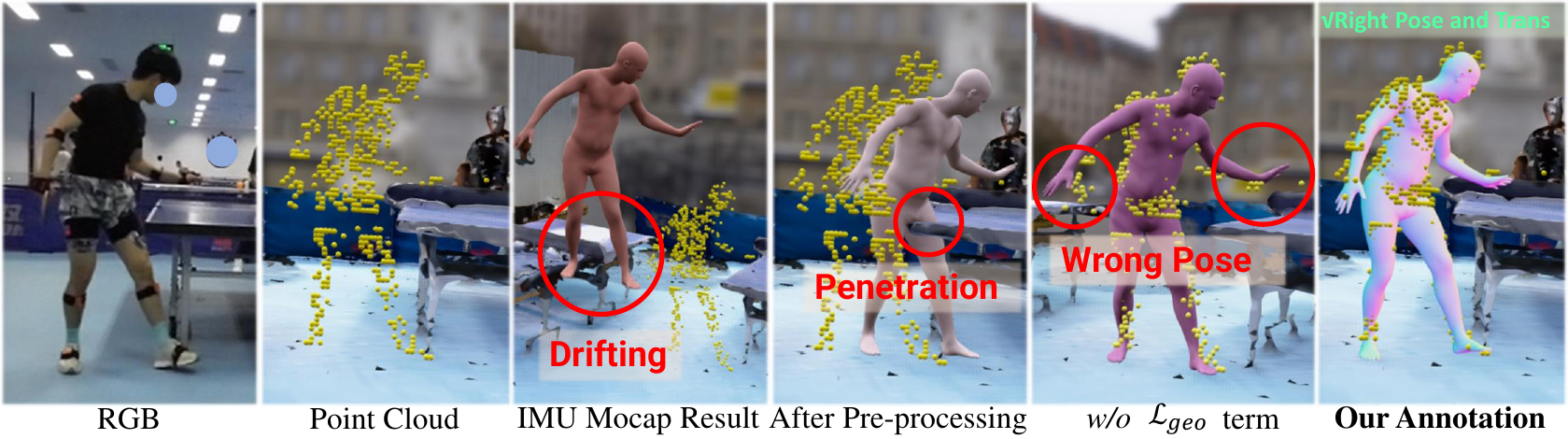}
     \vspace{-7mm}
     \caption{\textbf{Qualitative evaluation.} From left to right: RGB image, LiDAR point clouds, initial IMU motion capture result, 
     after pre-processing stage, after optimization without the $\mathcal{L}_{geo}$ loss, after optimization stage.}
     \label{fig:QualitativeRELI}
     \vspace{-5mm}
\end{figure}
We study the quality of the RELI11D dataset through qualitative and quantitative evaluations.


\PAR{Qualitative evaluation.}\cref{fig:QualitativeRELI} depicts a frame of the RELI11D dataset. The global translation of IMUs may be in-precise as it is floating in the air (column 3 of \cref{fig:QualitativeRELI}). Through the pre-processing stage, the quality of the dataset is improved, the position of the SMPL model is on the ground (column 4 of \cref{fig:QualitativeRELI}). However, it may cause penetration. In the consolidated optimization stage, RELI11D uses the contact aware loss to avoid penetration so that the quality is further improved. (column 5 of \cref{fig:QualitativeRELI}). Nevertheless, limb poses may be wrong. Through using the $\mathcal{L}_{geo}$ loss, we obtain the accurate poses and translations (the last column of \cref{fig:QualitativeRELI}).

\PAR{Quantitative evaluation.} To quantitatively evaluate the annotation quality of RELI11D, we select annotated motion sequences, and evaluate the performance of the consolidated optimization stage (in~\cref{Sec:Blending optimization}) by comparing the generated annotations against optimized annotations. \cref{tab:eva_annotation} depicts the error metrics for the annotations generated without/with the optimization stage in mean/max. The error metrics are small, which demonstrates the effectiveness of the annotation pipeline and the high quality of RELI11D.

In order to understand the impact of different constraints used in the consolidated optimization stage, we conduct a ablation study on 3 different losses: $\mathcal{L}_{contact}$, ${\mathcal{L}}_{smoo}$ and $\mathcal{L}_{geo}$. ~\cref{tab:eva_RELI} shows the error metrics of using different combinations of losses for two scenes. Without using any losses (row 1), the error metrics are the largest. If we remove any loss from the optimization stage (row 2 to 4), the error metrics increase, indicating all the losses are useful to improve the quality of our dataset. 



\begin{figure*}[tb]
    \vspace{-3mm}
    \centering
     \includegraphics[width=1\linewidth]{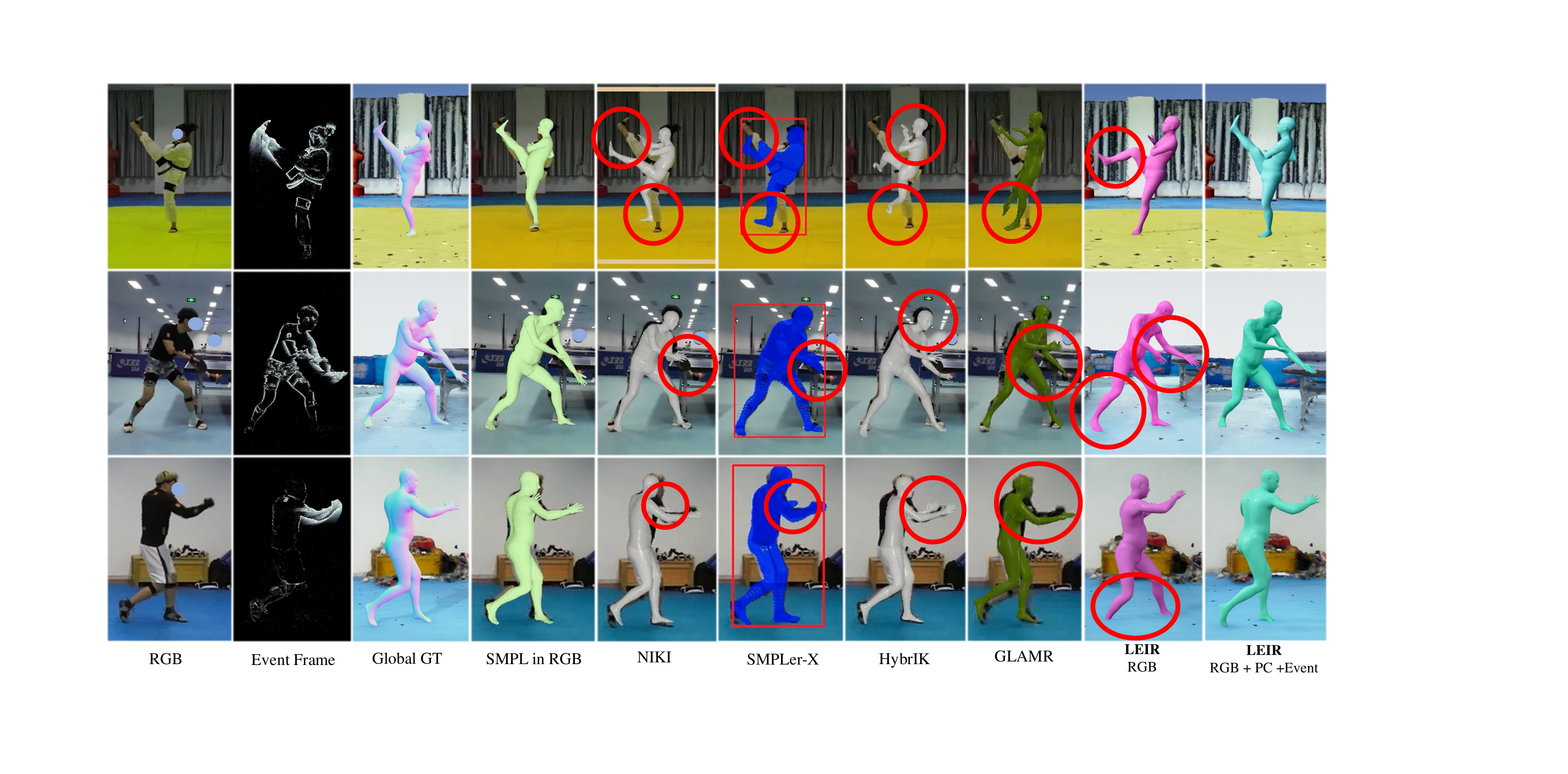}
     \vspace{-8mm}
     \caption{\textbf{Benchmark experiment and RELI11D demonstration.} Image-based methods (left col 5 to 9) can produce erroneous results due to occlusions and rapid movements (\textcolor[rgb]{1, 0, 0}{red circles}). Our multimodal method (first right col) performs best qualitatively by comparison.}
    \vspace{-4mm}
     \label{fig:QualitativeTask}
\end{figure*}

\subsection{Benchmark}\label{sec:tasks}
\vspace{-1mm}
The rich modalities provided in RELI11D allow us to conduct a systematic benchmark on HPE methods. We consider 5 HPE tasks: LiDAR-based, RGB-based, RGB+LiDAR-based, Event-based, and HPE with global trajectory. 

\begin{table}[tb]
	\centering

	\centering
	\resizebox{\linewidth}{!}{
    \begin{tabular}{c|cccccc}
    
     \toprule
     Input Modality & Method & ACCEL$\downarrow$ & MPJPE$\downarrow$ & PA-MPJPE$\downarrow$ & PVE$\downarrow$ & PCK0.3$\uparrow$ \\
     \midrule
     \multirow{3}*{LiDAR}  

     ~ & P4Tranformer~\cite{Fan2021Point4T} & 66.33 & 172.04 & 150.65 & 206.75 &  0.51 \\
     ~ & PCT~\cite{guo2021pct} & 59.19 & 144.40 & 116.99 & 174.33 & 0.67 \\
     ~ & LiDARCap~\cite{lidarcap}   & 54.42 & 144.51 & 106.20 & 176.98 & 0.67 \\
    \midrule
    \multirow{3}*{RGB} 
        & HybrIK~\cite{li2021hybrik}  & 58.39 & 249.34 & 163.91 & 255.98 & 0.53  \\
        & NIKI~\cite{li2023niki}  & 55.62 & 196.68 & 142.48 & 198.10 & 0.61 \\
        & SMPLer-X~\cite{SMPLERX}  & 50.15 & 171.97 & 128.02 & 185.83  & 0.66 \\
    \midrule
    
    \multirow{2}*{Event} & EventHPE~\cite{EventHPEICCV21} & - & 193.7 & 115.72 & 224.59 & 0.52 \\
        & EventPointPose~\cite{EventPointPose3DV22} & - & 16.24(2D) & 10.91(2D) & - & 0.69(2D) \\
    \midrule  

    \multirow{2}{1cm}{\centering RGB+ \\ LiDAR}

        & ImmFusion~\cite{chen2023immfusion} & 49.19 & 175.00 & 159.62 & 187.31 & 0.67 \\
        & FusionPose~\cite{cong2022weakly} & 44.89 & 136.15 & 110.19 & 166.94 & 0.75 \\
     \bottomrule
     \end{tabular}%
     }
     \vspace{-3mm}
     \caption{Comparison of SOTA HPE methods on different modals in RELI11D. Unit: $mm$}
     \vspace{-3mm}
	\label{tab:single_pose_estimation}
\end{table}
\begin{table}[!tb]
	\centering

	\centering
	\resizebox{\linewidth}{!}{
    \begin{tabular}{c|ccccccc}
    
     \toprule
     Input Modality & Method & ACCEL$\downarrow$ & MPJPE$\downarrow$ & PA-MPJPE$\downarrow$ & GMPJPE$\downarrow$ & T-Error$\downarrow$ & PCK0.3$\uparrow$ \\
    \midrule

     \multirow{2}{2cm}{\centering RGB} 
    & \centering GLAMR~\cite{GLAMR} & 47.83 & 202.66 & 179.59 & 495.40 & 590.46 & 0.65 \\
    
    & \centering TRACE~\cite{sun2023trace} & 50.09 & 197.41 & 165.18 & 488.91 & 581.67 & 0.68 \\
    
     \bottomrule
     \end{tabular}%
     }
     \vspace{-3mm}
     \caption{Comparison of SOTA 3D HPE methods \textbf{with global translation} in RELI11D. Unit: $mm$}
     \vspace{-6mm}
	\label{tab:benchmark_with_trans}
\end{table}
\vspace{-3mm}
\subsubsection{Human Pose Estimation (HPE)}
\vspace{-1mm}
We evaluate the performance of multiple state-of-the-art 3D HPE methods in RELI11D. Based on their input modality, we categorize these methods into four categories. For LiDAR-based input, LiDARCap~\cite{lidarcap}, P4Transformer~\cite{Fan2021Point4T}, and PCT~\cite{guo2021pct} are used. For RGB-based input, NIKI~\cite{li2023niki} and SMPLer-X~\cite{SMPLERX} are used. For event-based input, EventHPE~\cite{EventHPEICCV21} and EventPointPose~\cite{EventPointPose3DV22} are compared. For the RGB and LiDAR inputs, FusionPose~\cite{cong2022weakly} and ImmFusion~\cite{chen2023immfusion} are tested.


\cref{tab:single_pose_estimation} shows the HPE experimental results. We find that all the methods perform poorly on the RELI11D dataset, even for methods trained on multiple datasets. \cref{fig:QualitativeTask} shows that some methods do not work well for certain poses in the RELI11D dataset. The motions in taekwondo sports are rapid, their range is rarely seen in daily activities. As is depicted in the first row of \cref{fig:QualitativeTask}, all the RGB-based methods do not model the limbs well. For Table Tennies, besides modeling the rapid movement of hands and feet, their global position should be captured to avoid penetration. As shown in the second row of \cref{fig:QualitativeTask}, all the RGB methods have prediction artifacts. This indicates that RELI11D is a \emph{challenging dataset} for existing methods. 

As it is shown in \cref{tab:single_pose_estimation}, using two modalities (RGB+LiDAR) leads to better performance than using these two modalities separately. The rich modalities provided by RELI11D enable the comparison of different combinations of modality-based methods. 

\vspace{-3mm}
\subsubsection{Global HPE}
\vspace{-1mm}
Capturing certain actions, such as playing table tennis, requires the precise locations of humans. We evaluate two RGB-based global 3D pose estimation methods (GLAMR~\cite{GLAMR} and TRACE~\cite{sun2023trace}) on the RELI11D dataset. Their results are depicted in 
\cref{tab:benchmark_with_trans}, which shows that their performance is unsatisfactory. This suggests that despite the monocular camera methods can match local motions in 2D RGB well, they do not excel at 3D global motions.


\begin{table}[!tb]
	\centering

	\centering
	\resizebox{\linewidth}{!}{
    \begin{tabular}{c|cccccc}
    
     \toprule
     Input Modality &  ACCEL$\downarrow$ & MPJPE$\downarrow$ & PA-MPJPE$\downarrow$ & G-MPJPE$\downarrow$ & T-Error$\downarrow$ & PCK0.3$\uparrow$ \\
    \midrule 
    
    



    \centering LiDAR
    & \large{31.26} & \large{59.93} & \large{48.23} & \large{125.77} & \large{195.72} & \large{0.89} \\

    \centering RGB
    & \large{28.43} & \large{62.71} & \large{54.11} & \large{557.81} & \large{710.84} & \large{0.88} \\

    \centering Event
    & \large{34.45}& \large{107.78} & \large{83.64} & \large{605.45} & \large{743.71} & \large{0.59} \\

    \midrule
     \multirow{2}{2cm}{\centering LiDAR+ \\ \centering RGB} 
    & \multirow{2}*{\large{27.07}} & \multirow{2}*{\large{55.36}} & \multirow{2}*{\large{45.72}} & \multirow{2}*{\large{122.32}} & \multirow{2}*{\large{168.61}} & \multirow{2}*{\large{0.90}} \\
    &  &  &  &  &   & \\

     \multirow{2}{2cm}{\centering LiDAR+ \\ \centering Event} 
    & \multirow{2}*{\large{25.41}} & \multirow{2}*{\large{57.79}} & \multirow{2}*{\large{46.70}} & \multirow{2}*{\large{123.75}} & \multirow{2}*{\large{178.97}} & \multirow{2}*{\large{0.89}} \\
    &  &  &  &  &   & \\

    \midrule

    \multirow{2}{2cm}{\centering LiDAR+ \\ \centering RGB+Event} 
        & \multirow{2}*{\large{\textbf{23.90}}} & \multirow{2}*{\large{\textbf{49.19}}} & \multirow{2}*{\large{\textbf{40.87}}} & \multirow{2}*{\large{\textbf{115.36}}} & \multirow{2}*{\large{\textbf{146.13}}} & \multirow{2}*{\large{\textbf{0.92}}} \\
        &  &  &  &  &   &  \\
     \bottomrule
     \end{tabular}%
     }
     \vspace{-3mm}
     \caption{The performance of LEIR input with different modalities based on RELI11D. Unit: $mm$}
     \vspace{-3mm}
	\label{tab:baseline_table}
\end{table}
\begin{table}[tb]
	\centering

	\centering
	\resizebox{\linewidth}{!}{
    \begin{tabular}{c|cccccc}
    
     \toprule
     Input Modality & Method & ACCEL$\downarrow$ & MPJPE$\downarrow$ & PA-MPJPE$\downarrow$ & PVE$\downarrow$ & PCK0.3$\uparrow$ \\
    \midrule 
    
     \multirow{2}{2cm}{\centering LiDAR} 
    & LiDARCap~\cite{lidarcap} & \large{45.89} & \large{80.08} & \large{67.50} & \large{102.24} & \large{0.85} \\
    & \textbf{LEIR(Ours)} & \large{\textbf{45.60}} & \large{\textbf{79.00}} & \large{\textbf{67.45}} & \large{\textbf{100.87}} & \large{\textbf{0.85}} \\
    
    
    \midrule

    \multirow{3}{2cm}{\centering LiDAR+ \\ \centering RGB} 
    & ImmFusion~\cite{chen2023immfusion} & \large{46.45} & \large{96.93} & \large{81.16} & \large{107.29} & \large{0.75} \\
    & FusionPose~\cite{cong2022weakly} & \large{\textbf{44.51}} & \large{78.18} & \large{66.70} & \large{99.66} & \large{0.85} \\
    & \textbf{LEIR(Ours)} & \large{44.52} & \large{\textbf{75.09}} & \large{\textbf{62.94}} & \large{\textbf{95.96}} & \large{\textbf{0.87}} \\

     \bottomrule
     \end{tabular}%
     }
     \vspace{-3mm}
     \caption{Performance evaluation of LEIR in the LiDARHuman26M dataset~\cite{lidarcap}. Unit: $mm$}
     \vspace{-5mm}
	\label{tab:baseline_LiDARHuman26M}
\end{table}

\subsection{Baseline Evaluation}
\label{sec:Baseline Evaluation}
We evaluate the proposed baseline, LEIR, based on the RELI11D and the LiDARHuman26M~\cite{lidarcap} datasets, with different combinations of modalities. 

\PAR{LEIR on RELI11D.} \cref{tab:baseline_table} presents the results of LEIR with different combinations of modalities as input. For single modality (rows 2 to 4), LiDAR-based input achieves the best performance due to its ability to provide detailed geometric information. RGB-based input achieves the second-best performance, benefiting from its appearance information. Event-based input yields the worst results. As the number of modalities increases, the performance of LEIR improves. This emphasizes that correctly fusing the information of each modality feature makes the method robust. When combining LiDAR with RGB/Event inputs (rows 5 and 6), LEIR outperforms the use of LiDAR alone. For HPE, the best performance is achieved when utilizing all the modalities (LiDAR+RGB+Event), and it is significantly better than all the studied state-of-the-art methods. 

Regarding global HPE, compared with the two RGB-based methods (shown in \cref{tab:benchmark_with_trans}), the results shown in \cref{tab:baseline_table} demonstrate that using the LiDAR modality is necessary for global pose estimation. Combining all three modalities can achieve the best global HPE results.

These results highlight the effective utilization of information from different modalities in LEIR, and they are visualized in \cref{fig:QualitativeTask}, intuitively showing that LEIR (RGB+PC+Events) works well on the RELI11D dataset.

We conduct a study on another dataset LiDARHuman26M~\cite{lidarcap}, which contains both RGB and LiDAR modalities. In this experiment, we train all the methods from scratch based on the methodology described in \cite{lidarcap} and follow the same evaluation as \cite{lidarcap}. As is shown in \cref{tab:baseline_LiDARHuman26M}, when considering LiDAR input alone, LEIR demonstrates a slight improvement over LiDARCap. When combining LiDAR and RGB inputs, LEIR out-performs ImmFusion ~\cite{chen2023immfusion} and FusionPose ~\cite{cong2022weakly}. This indicates that our proposed method, LEIR, performs well on other dataset.

\PAR{Global trajectory prediction.} The T-Error measures the translation error is depicted in ~\cref{tab:baseline_table}. And the predicted trajectory is plotted in Supp Fig.3. It shows that incorporating the LiDAR point clouds with global trajectory information improves the global motion indicators (low T-Error, similarity between the curve and ground truth). This observation indicates a promising trend of multimodal methods that fuse global information. 
\section{Conclusion}
Different motion sensors have distinct characteristics (e.g., geometry) that excel at capturing challenging motions (e.g., complex and fast motion). We introduce RELI11D, the first human motion dataset with the LiDAR, RGB, IMU, and Event modalities for a holistic understanding of human motions. It records the motions of 10 actors performing 5 sports in 7 different scenes. The rich annotations in RELI11D enable benchmarking a series of HPE tasks. We demonstrate that RELI11D is challenging due to its fast and complex motions. To address this challenge, we propose LEIR, a multimodal HPE baseline that utilizes the LiDAR points cloud, event streams, and RGB videos through cross-attention strategy. We show through extensive experiments that LEIR can obtain the most competitive results.

\PAR{Acknowledgements.} This work was partially supported by the National Natural Science Foundation of China (No.62171393), by
the Fundamental Research Funds for the Central Universities (No.20720230033, No.20720220064), by PDL (2022-PDL-12).




{\small
\bibliographystyle{ieee_fullname}
\bibliography{egbib}
}

\end{document}